\documentclass[conference]{IEEEtran}
\IEEEoverridecommandlockouts

\usepackage{graphicx}
\usepackage{amsmath}
\usepackage{amssymb}
\usepackage{booktabs}
\usepackage{textcomp}
\usepackage{algorithmic}
\usepackage{float}
\usepackage{xcolor,colortbl}
\usepackage{adjustbox}
\usepackage{cuted}
\usepackage{caption}
\usepackage{scalerel}
\usepackage[pagebackref,breaklinks,colorlinks]{hyperref}
\usepackage{amssymb} 
\usepackage{multicol}

\definecolor{Gray}{gray}{0.90}
\title{EDADepth: Enhanced Data Augmentation for Monocular Depth Estimation}

\begin{document}

\author{Nischal Khanal and Shivanand Venkanna Sheshappanavar\\
Geometric Intelligence Research Lab, University of Wyoming, USA.\\}
\twocolumn[{
\renewcommand\twocolumn[1][]{#1}
\maketitle
\begin{center}
    \captionsetup{type=figure}
    \includegraphics[width=0.99\linewidth]{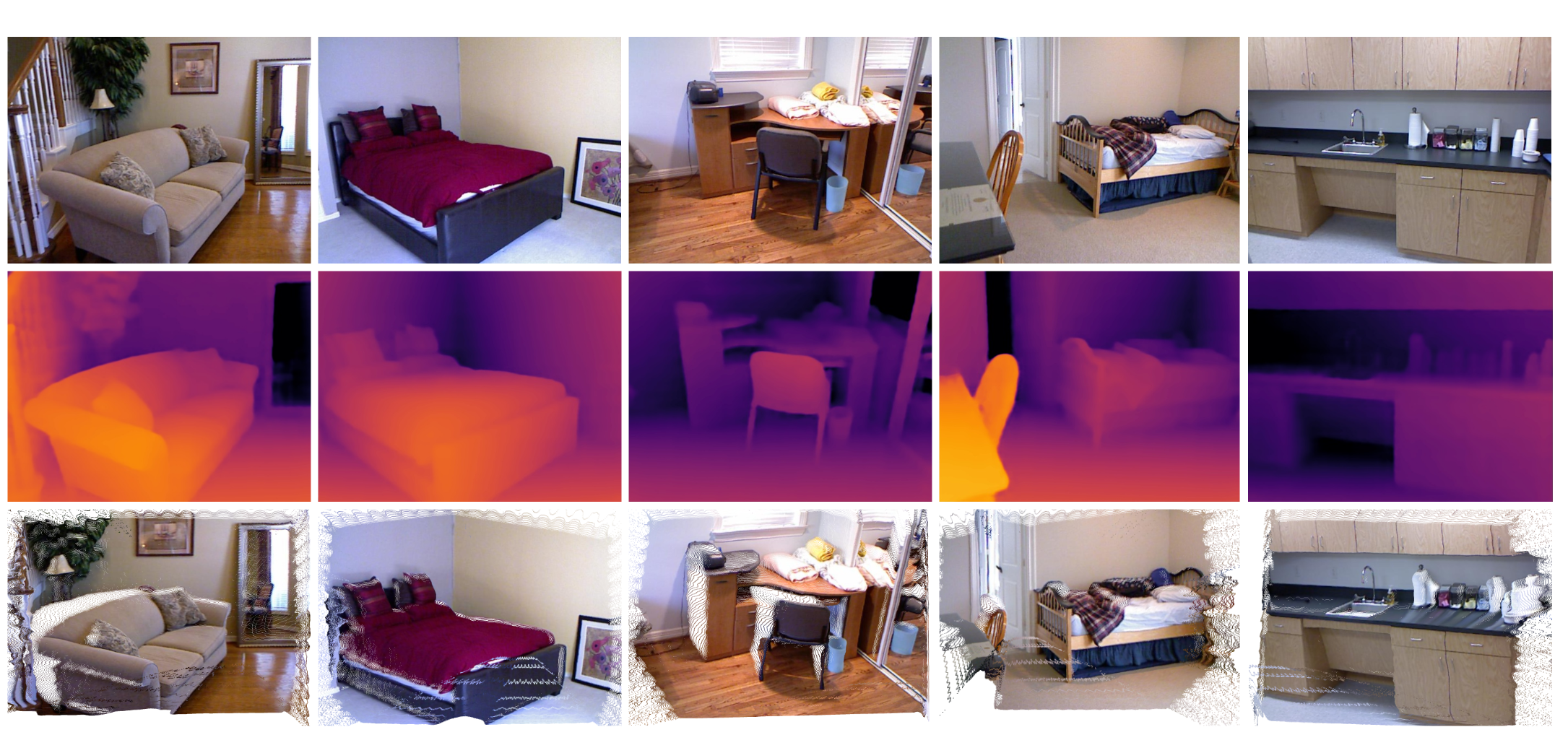}
    \captionof{figure}{Our EDADepth model takes single image \textbf{(top)} and estimates depth using a pre-trained U-Net model. It uses the BEiT semantic segmentation model to extract context for the generation of depth maps \textbf{ (Middle)}. 3D point cloud (\textbf{Bottom}) is constructed from the estimated depth map and the respective input RGB image.}
    \label{fig:abstract}
 \end{center}
}]


\begin{abstract}

Due to their text-to-image synthesis feature, diffusion models have recently seen a rise in visual perception tasks, such as depth estimation. The lack of good-quality datasets makes the extraction of a fine-grain semantic context challenging for the diffusion models. The semantic context with fewer details further worsens the process of creating effective text embeddings that will be used as input for diffusion models. In this paper, we propose a novel EDADepth, an enhanced data augmentation method to estimate monocular depth without using additional training data. We use Swin2SR, a super-resolution model, to enhance the quality of input images. We employ the BEiT pre-trained semantic segmentation model to better extract text embeddings. We use the BLIP-2 tokenizer to generate tokens from these text embeddings. The novelty of our approach is the introduction of Swin2SR, the BEiT model, and the BLIP-2 tokenizer in the diffusion-based pipeline for the monocular depth estimation. Our model achieves state-of-the-art results (SOTA) on the $\delta_3$ metric on the NYUv2 and KITTI datasets. It also achieves results comparable to those of the SOTA models in the RMSE and REL metrics. Finally, we also show improvements in the visualization of the estimated depth compared to the SOTA diffusion-based monocular depth estimation models—code repository: \url{https://github.com/edadepthmde/EDADepth_ICMLA}. 
\end{abstract}
\section{Introduction}

Depth estimation is an essential task in computer vision that measures the distance of each pixel relative to a camera. Depth is necessary for operations such as 3D reconstruction~\cite{zollhofer2018state} (refer \ref{fig:abstract}), scene understanding~\cite{zhou2016places} and recognition~\cite{sheshappanavar2024benchmark}. One of the types of depth estimation is Monocular Depth Estimation (MDE)~\cite{ming2021deep}. MDE is a task that estimates the depth of an object using a single-view image. Since single-view images do not have epipolar geometry~\cite{zhang1998determining}, it is challenging to determine the depth of each pixel. Traditional methods for depth estimation used monocular cues \cite{saxena2007depth} and shading \cite{shadingmde}. However, such methods faced challenges, such as the varying image lighting and the need for precise camera calibration. Such limitations suggested a need for a technique to estimate depth value based on per-pixel regression\cite{mdereview2}, a task commonly used in deep learning ~\cite{ming2021deep}. Hence, deep learning methods have emerged as a reliable solution for depth estimation. 

One of the most popular deep learning methods used in computer vision is Transformers~\cite{vaswani2017attention}. Transformers employ self-attention mechanisms, making them a good choice for capturing long-range dependencies~\cite{ZUO2022109552}. Long-range dependencies are significant in MDE, as they capture contextual information from various regions in a single image. Hence, Transformers have successfully been applied to estimate monocular depth. Furthermore, transformers effectively create generative models such as the diffusion model~\cite{rombach2021highresolution}. Diffusion models have been widely used for text-to-image generation and image denoising. Additionally, data augmentation techniques have significantly improved the performance of Transformers on various datasets \cite{steiner2022trainvitdataaugmentation}. In this paper, we introduce a diffusion-based model called EDADepth, an enhanced data augmentation-based monocular depth estimation.

\begin{figure}[htbp] 
    \centering
    \includegraphics[width=1.0\linewidth]{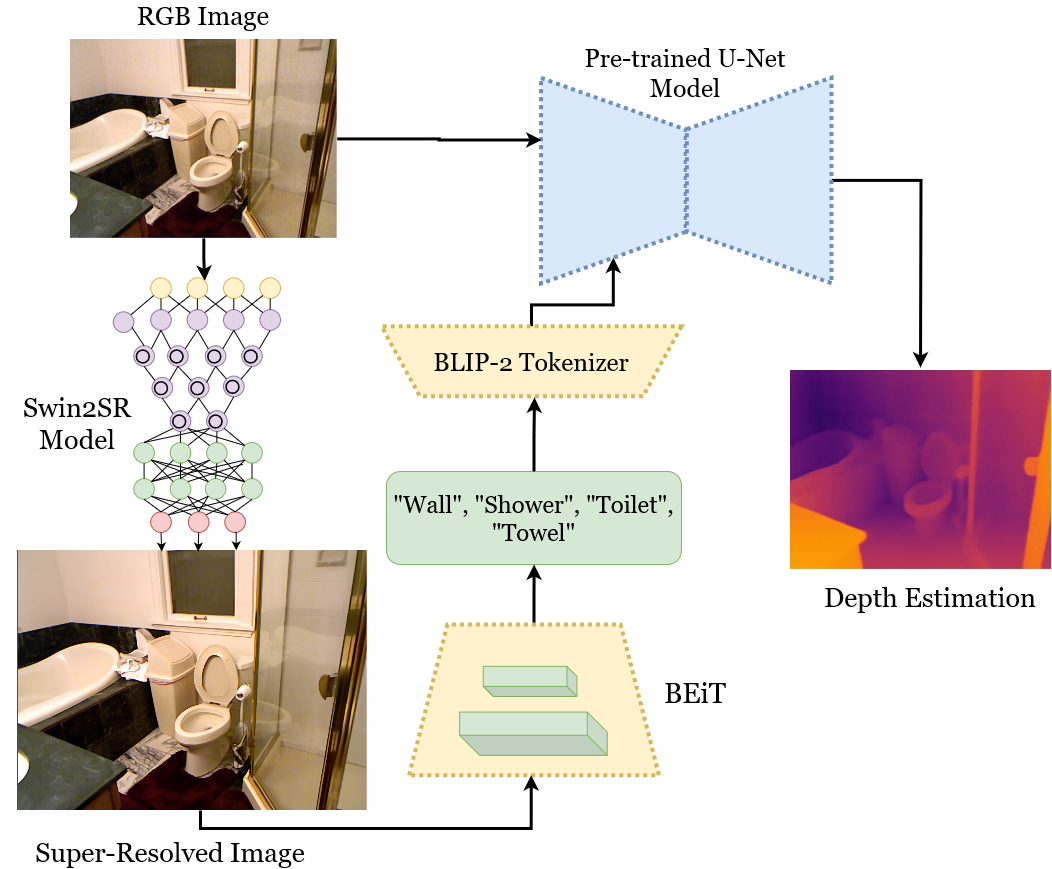}
    \caption{In EDADepth, the raw RGB input image is enhanced using a Swin2SR model. BEiT model extracts detailed semantic context from the enhanced image and passes it to a BLIP-2 tokenizer for tokens. These text embedding tokens are fed to a pre-trained U-Net model to estimate depth.}
    \label{concept}
\end{figure}

In EDADepth, we created a diffusion-based pipeline that does not use extra training data, following the footsteps of ECoDepth\cite{patni2024ecodepth}. Our pipeline enhances the input image through data augmentation. The original indoor NYU--Depth V2~\cite{Silberman:ECCV12} dataset with low image quality has been fed to a pre-trained Swin2SR model~\cite{conde2022swin2srswinv2transformercompressed} for obtaining the enhanced dataset. From the input data, to extract text-embeddings, pre-trained ViT~\cite{wu2020visual} and CLIP~\cite{clipmodel} models are widely used. Our model introduces a novel idea of using a pre-trained BEiT semantic segmentation model~\cite{bao2022beitbertpretrainingimage} for extracting detailed text embeddings as summarized in Figure \ref{concept}. For our experiments and visualization, we used both the indoor (NYU--Depth V2) \cite{Silberman:ECCV12} and outdoor (KITTI) ~\cite{Geiger2012CVPR} datasets. Following recent works \cite{patni2024ecodepth}, \cite{zeng2024wordepthvariationallanguageprior}, the evaluation metrics are absolute relative error (REL) and root mean squared error (RMSE), the average error $log_{10}$ between the predicted depth and the actual depth and threshold accuracy $\delta_{n}$. 

The key contributions of this work are three-fold:
\begin{itemize}
    \item We propose a novel method to enhance the input images to improve the estimated depth map. The enhanced input is used for semantic context extraction. 
    \item We adopt a BEiT semantic segmentation model to extract the semantic context for creating text embeddings. We employ the BLIP-2 tokenizer as a novel way to create text embedding tokens from the extracted semantic context.
    \item We provide both qualitative and quantitative evaluations on two popular datasets NYUv2\cite{Silberman:ECCV12} and KITTI ~\cite{Geiger2012CVPR} to demonstrate the effectiveness of our pipeline.  
\end{itemize}

\section{Related Works}
\subsection{Monocular Depth Estimation}
Over the last decade, several methods~\cite{eigen2014depthmappredictionsingle,fu2018deepordinalregressionnetwork,yin2018geonetunsupervisedlearningdense,lee2021bigsmallmultiscalelocal,Farooq_Bhat_2021,ranftl2021visiontransformersdenseprediction,yuan2022newcrfsneuralwindow} have been proposed addressing Monocular Depth Estimation (MDE). MDE using supervised~\cite{supervised} and self-supervised learning~\cite{Gasperini_2023_ICCV} are among the recent works. The first MDE challenge organized at WACV 2023 showcased the work of Spencer et al.~\cite{Spencer_2023_WACV}. Several participants at the challenge outperformed the baseline on the SYNS-Patches~\cite{Spencer2022} dataset. Few among the teams that implemented self-supervised learning models were team OPDAI whose MDE model was based on ConvNeXt-B~\cite{liu2022convnet} and HRDepth~\cite{lyu2020hrdepth}  models. Team z.suri based their MDE model on ConvNeXt~\cite{liu2022convnet} and DiffNet~\cite{park2021diffnet} models and team MonoViT's~\cite{Zhao_2022} model was based on MPViT \cite{lee2021mpvitmultipathvisiontransformer} model. ZoEDepth~\cite{bhat2023zoedepth} introduced a generalized and robust method for MDE using zero-shot transfer knowledge. 

\subsection{Diffusion-based MDE models}

Recently, diffusion model~\cite{rombach2021highresolution} has attracted more notable advances in estimating depth \cite{zhao2023unleashing,kondapaneni2024tadp,ke2023repurposing,wan2023harnessing,ji2023ddpdiffusionmodeldense} because of its pre-trained features. Due to their nature of intentionally adding noise into the data during the forward process and trying to restore the original data during the reverse process, they have been extensively used to extract high-level features for MDE. 

\begin{figure*}[ht] 
    \centering
    \vspace{-6pt}
    \includegraphics[width=\linewidth, keepaspectratio=TRUE]{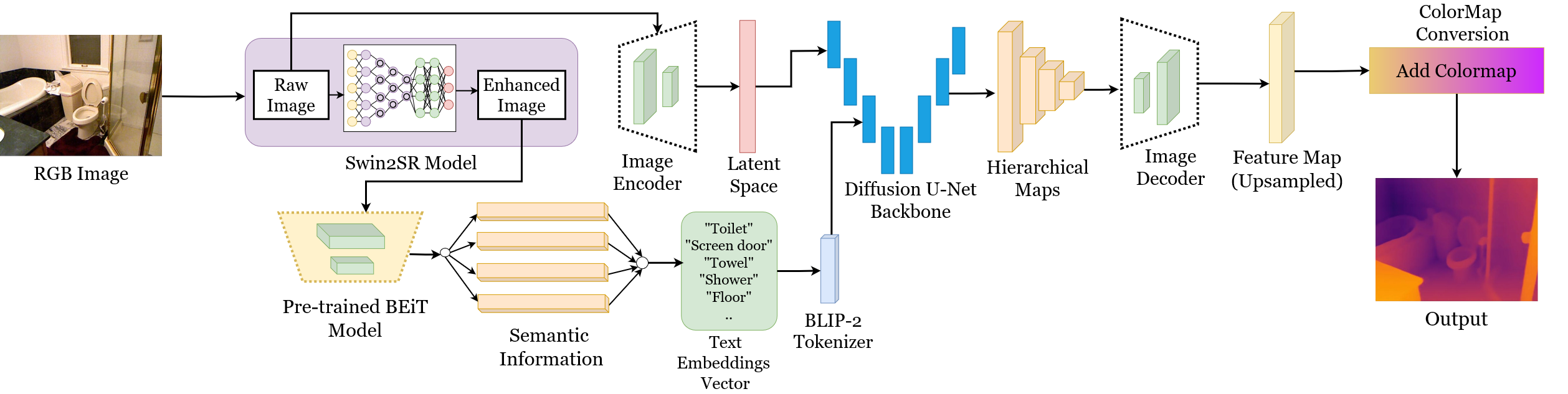}
    \caption{\textbf{EDADepth model framework}. The architecture integrates a Swin2SR model to process raw RGB inputs, producing enhanced images for the text embedding module. It utilizes the BEiT semantic segmentation model for a segmentation-based, self-supervised text embedding process that generates a vector of text embeddings. These vectors are then fed into the U-Net model via the BLIP-2 tokenizer. The model follows a forward-reverse denoising process to generate an estimated depth map. }
    \label{architecture}
\end{figure*}

VPD~\cite{zhao2023unleashing} uses the U-Net architecture for depth estimation and reference segmentation. EVP~\cite{lavreniuk2023evp} and MetaPrompts~\cite{wan2023harnessing} enhance the existing VPD model by adding layers to create effective text embeddings. Recently, ECoDepth~\cite{patni2024ecodepth} introduced the concept of using embeddings from a pre-trained ViT for detailed semantic information extraction. Existing MDE models \cite{zhao2023unleashing,patni2024ecodepth,wan2023harnessing,lavreniuk2023evp,kondapaneni2024tadp} use CLIP \cite{clipmodel} text tokenizer for generating text embedding tokens from semantic context. Our approach differs from traditional image-to-text caption generators such as CLIP \cite{clipmodel} used in VPD \cite{zhao2023unleashing}, providing a more informative and precise representation of the input images. Likewise, PatchFusion~\cite{li2024patchfusion} was the first to enhance low-quality input dataset as a data augmentation step in the MDE pipeline.  In the same direction, we for the first time, use the Swin2SR model to enhance the input dataset. To effectively extract the semantic context from the enhanced dataset, we introduce a novel idea of using the BEiT semantic segmentation~\cite{bao2022beitbertpretrainingimage} model. Additionally, to extract tokens from the text embeddings effectively, we propose applying BLIP-2 \cite{li2023blip2} tokenizer.

\section{Methods}
\subsection{Diffusion Models Overview}

Diffusion models~\cite{rombach2021highresolution} are generative models implemented by adding noise to the input and aiming to reconstruct the original input by learning the reverse denoising process. The diffusion model implemented in this project is stable diffusion, a text-to-image latent diffusion model ~\cite{rombach2021highresolution}. The Stable Diffusion model consists of four key components: Encoder (E), Conditional denoising auto-encoder ($\epsilon_{0}$), Language encoder ($\tau_{\theta}$), and decoder (D). The diffusion process is modeled as follows:
\begin{equation}
z_t \sim \mathcal{N}(\sqrt{\alpha_t}z_{t-1},(1-\alpha_t)\boldsymbol{I})
\label{eq:1}
\end{equation}
where $z_t$ is random variable at time $t$, $\alpha_t$ is fixed coefficient representing the noise schedule, and $\mathcal{N}(z,\mu,\sigma)$ represents the normal distribution. 

The encoder (E) and decoder (D) are trained before the $\epsilon_{0}$, such that  $D(E(x))=\hat{x} \approx x$. $\epsilon_{0}$ is implemented using U-Net as a pre-trained forward process (using the LAION-5B dataset~\cite{schuhmann2022laion5b}) and we train the reverse process for depth estimation. $\epsilon_{0}$ of the latent diffusion model is trained to minimize the loss given by:
\begin{equation}
    L_{LDM} := \mathbb{E}_{E(x),y,\epsilon\sim\mathcal{N}(0,1),t} ||\epsilon-\epsilon_{\theta}(z_t,t,\tau_\theta(y))||^2_2
    \label{eq:2}
\end{equation}
where $z_t$ is calculated using Equation \ref{eq:1}. 

From equation \ref{eq:1}, we know that the diffusion model is processed as Markov, making it a regression problem, which can be used to model the distribution $p(y|x)$, where \textit{y} is the output depth and \textit{ x} is its corresponding input image. Since we already have a pre-trained model from stable diffusion, the model $\epsilon_0$ can be used to predict the density function gradient, $\nabla_{z_t}log p(z_t|C)$. The distribution $p(y|x)$ can be further modeled as:
\begin{equation}
    p_{\lambda}(y|x,\mathcal{T})=p_{\lambda_{4}}(y|z_0)p_{\lambda_{3}}(z_0|z_t,\mathcal{C})p_{\lambda_{2}}(z_t|x)p_{\lambda_{1}}(\mathcal{C},x)
    \label{eq:3}
\end{equation}
 where, $p_{\lambda_{1}}(\mathcal{C},x)= p(\mathcal{C}|\mathcal{T})p(\mathcal{T}|x)$. 

$\mathcal{T}$ is used to denote the textual embeddings obtained from the BEiT model (discussed in the next section). The pre-trained transformer model implements the distribution $p(\mathcal{T}|x)$, and through the learnable embeddings from the BEiT model, $p(\mathcal{C}|\mathcal{T})$ is implemented. The distribution $p(z_0|x)$ is implemented using the encoder (\textit{E}). Likewise, the distribution $p(z_0|z_t,\mathcal{C})$ is implemented via the U-Net model ~\cite{ronneberger2015unet} to extract hierarchical feature maps. Finally, the distribution $p(y|z_0)$ generates the depth map from the hierarchical feature maps. 

\begin{figure}[ht] 
    \centering
    \includegraphics[width=1.0\linewidth]{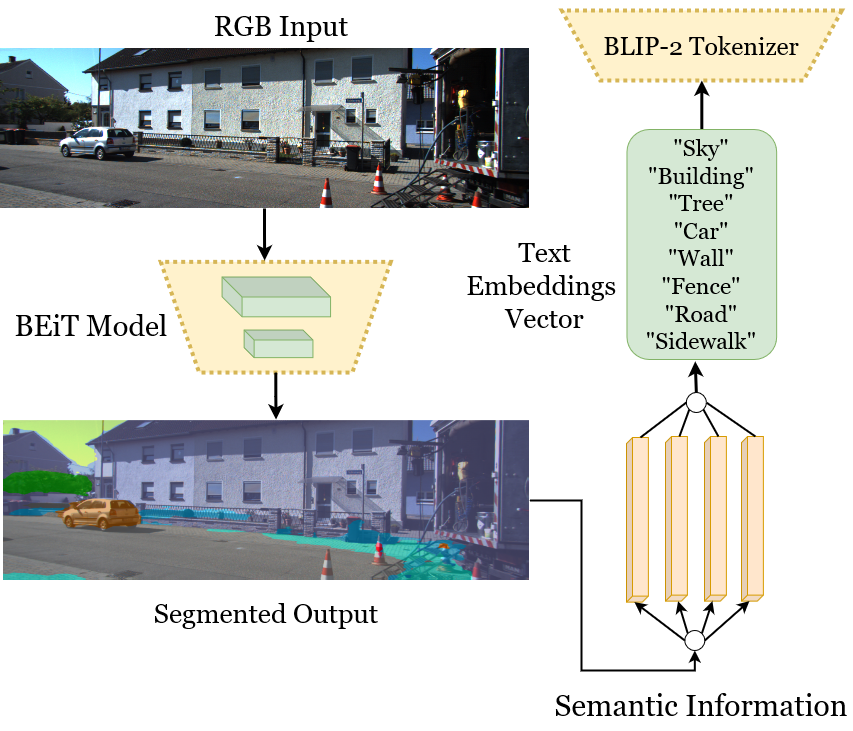}
    \caption{Text-embedding extraction using BEiT model.}
    \label{swin}
\end{figure}

\begin{figure*}[!htbp]
\centering
\includegraphics[width=1.0\linewidth]{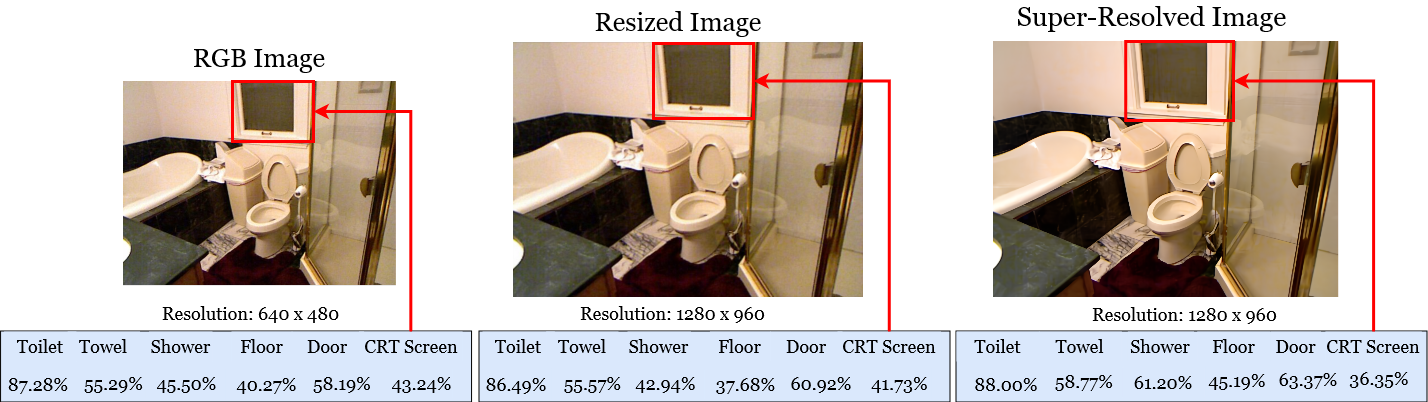}
\caption{Probabilities of the predicted semantic classes for the original, the resized, and the super-resolved images.}
\label{enhancement}
\end{figure*}

\subsection{BEiT model for Text-Embeddings}
We adopt the BEiT model, a self-supervised semantic segmentation model trained on ImageNet-21K from the ImageNet dataset~\cite{deng2009imagenet} and fine-tuned on the ADE20K dataset ~\cite{deng2009imagenet} to extract semantic context from enhanced input images. Figure \ref{swin} describes the image semantic segmentation pipeline of the BEiT model that extracts the semantic context of the image into a 150-dimensional logit vector. The semantic context is fed to a multilayered perceptron equipped with GELU\cite{hendrycks2023gaussianerrorlinearunits} to generate text embeddings (100-dimensional logit vectors). These vectors are now passed to the BLIP-2 tokenizer~\cite{li2023blip2}. Our model then performs a forward-reverse denoising process to extract high-level knowledge based on the learnable embeddings.

\subsection{Enhanced Data Augmentation}
Depth estimation models are sensitive to low-quality inputs~\cite{li2024patchfusion}, which can lead to loss of features. When a low-quality input image is supplied to the BEiT text embedding model, it results in incomplete semantic information or knowledge and improper estimation of depth for various objects. Hence, we propose an enhanced data augmentation through a pre-trained Swin2SR model~\cite{conde2022swin2srswinv2transformercompressed} to enhance the quality of the input image. Figure~\ref{enhancement} compares the probability of BEiT predicted semantic classes between the original images resized using bilinear interpolation and the superresolved images. There is a noticeable improvement in the probability of the super-resolved image. The model misclassifies the "windowpane" class as the "CRT screen" class in the red-annotated portions of the images, likely because of their similar visual appearances. However, the predictive probability of the superresolved image is lower than that of the original image, indicating that the superresolved input enhances the accuracy of semantic classification while reducing false positives. In addition, the probabilities for many components of the resized image are lower than those of the original. 

\subsection{Overall EDADepth Architecture}
Figure~\ref{architecture} shows the overall architecture of EDADepth model.

\noindent\textbf{Input:} As shown in figure \ref{architecture}, the input RGB image is fed to an image encoder for conversion into a latent space ~\cite{liu2019latent}, which is then processed through the denoising U-Net. The same image is also provided to the Swin2SR model, which enhances the image by super-resolving it. The enhanced image is passed through the pre-trained BEiT model for semantic context extraction. This step transforms the input image into a sequence of visual tokens ~\cite{bao2022beitbertpretrainingimage} for semantic contextual information extraction to segment the image. This contextual information, after linear transformation  ~\cite{bao2023contextual} is passed through the U-Net diffusion backbone (stable diffusion).

\noindent\textbf{Stable Diffusion:} The stable diffusion model enables image diffusion into the latent space to learn latent embedding through a variational autoencoder (VAE)~\cite{doersch2021tutorial}. The VAE transforms the input image into latent space for the U-Net model which considers different features of the input image in various dimensions. Likewise, learnable text embeddings are fed into the U-Net. Based on the image denoising process and the text-to-image generation process from text embeddings in the U-Net model,  multiscale hierarchical feature maps~\cite{kavukcuoglu2010learning} are generated and sent to the upsampling decoder~\cite{ren2023pixel}. 
    
\noindent\textbf{Decoder:} The Decoder~\cite{ren2023pixel} is designed to perform convolution-deconvolution~\cite{chen2017low} to upsample the feature maps. The decoder has a regression model with two convolution layers that generate the depth map from the feature maps. This depth map is colorized to better visualize the metric depth estimate of the RGB input image. 

\begin{table*}[htbp]
    \centering
    \caption{Comparison of recent models on the NYUv2 Dataset. The recent models are categorized into non-diffusion-based and diffusion-based monocular depth estimation (MDE) models. Diffusion-based MDE models are further divided into those trained with extra training data (ETD) and those without. \textbf{Bold metrics} indicate SOTA performance, and \textit{italic metrics} indicate the second-best performance. The row with a light gray fill represents the performance of our model.}
\begin{tabular}{ c|c|c|c|c|c|c|c|c }
        \hline
        \textbf{Method} & 
        \textbf{Venue} & \textbf{RMSE$\downarrow$} & \textbf{REL$\downarrow$} & \textbf{$log_{10}$$\downarrow$} & \textbf{$\delta_{1}$$\uparrow$} & \textbf{$\delta_{2}$$\uparrow$} & \textbf{$\delta_{3}$$\uparrow$} &
        \textbf{extra training data}\\
        \hline
        \multicolumn{9}{c}{\textit{Non-Diffusion-Based}}\\
                \hline
        Eigen et al. \cite{eigen2014depthmappredictionsingle}  & NIPS'14        & 0.641 & 0.158 & - & 0.769  & 0.950  & 0.988 & $\times$ \\
        DORN \cite{fu2018deepordinalregressionnetwork}  &CVPR'18         & 0.509 & 0.115 & 0.051 & 0.828  & 0.965  & 0.992 & $\times$\\
        GeoNet \cite{yin2018geonetunsupervisedlearningdense}  & TPAMI'20        & 0.569 & 0.128 & 0.057 & 0.834  & 0.960  & 0.990 & $\times$\\
        SharpNet \cite{ramamonjisoa2019sharpnetfastaccuraterecovery}  & ICCVW'21        & 0.502 & 0.139 & 0.047 & 0.836  & 0.966  & 0.993 & $\times$\\
        Yin et al.\cite{yin2020learningrecover3dscene}   & CVPR'21 & 0.416 & 0.108 & 0.048 & 0.875  & 0.976  & 0.994 & $\times$\\
        BTS\cite{lee2021bigsmallmultiscalelocal}  &Arxiv'19         & 0.392 & 0.110 & 0.047 & 0.885  & 0.978  & 0.994 & $\times$\\
        ASN \cite{long2021adaptivesurfacenormalconstraint}  & ICCV'21        & 0.377 & 0.101 & 0.044 & 0.890  & 0.982  & 0.996 & $\times$\\
        TransDepth \cite{yang2021transformerbasedattentionnetworkscontinuous}  &ICCV'21         & 0.365 & 0.106 & 0.045 & 0.900  & 0.983  & 0.996 & $\times$\\
        AdaBins\cite{Farooq_Bhat_2021}   & CVPR'21        & 0.364 & 0.103 & 0.044 & 0.903  & 0.984  & 0.997 & $\times$\\
        DPT \cite{ranftl2021visiontransformersdenseprediction}  & ICCV'21        & 0.357 & 0.110 & 0.045 & 0.904  & 0.988 & 0.998 & \checkmark\\
        P3Depth\cite{patil2022p3depthmonoculardepthestimation}   & CVPR'22 & 0.356 & 0.104 & 0.043 & 0.898 & 0.981 & 0.996 & $\times$ \\
        NeWCRFs\cite{yuan2022newcrfsneuralwindow}  & CVPR'22        & 0.334 & 0.095 & 0.041 & 0.922    & 0.992 & 0.998 & $\times$\\
        Localbins \cite{bhat2022localbinsimprovingdepthestimation}  &ECCV'22         & 0.357 & 0.099 & 0.042 & 0.907  & 0.987 & 0.998 & $\times$\\
        DepthFormer\cite{Li_2023}  &ArXiv'22         & 0.329 & 0.094 & 0.040 & 0.923  & 0.989 & 0.997 & $\times$\\
        PixelFormer\cite{agarwal2022attentionattentioneverywheremonocular}  &  WACV'23       & 0.322 & 0.090 & 0.039 & 0.929  & 0.991  & 0.998 & $\times$\\  
        WorDepth \cite{zeng2024wordepthvariationallanguageprior}      & CVPR'24& 0.317 & 0.088 & 0.038 & 0.932  & 0.992  & 0.998 & $\times$\\
        MIM \cite{xie2022revealingdarksecretsmasked}  &   CVPR'23      & 0.287 & 0.083 & 0.035 & 0.949  & 0.994  & 0.999 & $\times$\\
        ZoeDepth \cite{bhat2023zoedepth}  &ArXiv'23& 0.270 & 0.075 & 0.032 & 0.955  & 0.995  & 0.999 & \checkmark\\

                \hline
        \multicolumn{9}{c}{\textit{Diffusion-Based (with extra training data)}}\\
                \hline
                
        VPD \cite{zhao2023unleashing}        &ICCV'23 & 0.254 & 0.069 & 0.030 & 0.964  & 0.995  & 0.999 & \checkmark\\
        TADP   \cite{kondapaneni2024tadp}     & CVPR'24& 0.225 & 0.062 & 0.027 & 0.976  & 0.997  & 0.999 & \checkmark\\
        Marigold \cite{ke2023repurposing} &CVPR'24 &0.224 &\textbf{ 0.055} & \textbf{0.024}    & 0.964  & 0.991  & 0.998& \checkmark\\
        MetaPrompts \cite{wan2023harnessing} &ArXiv'23 &0.223 & 0.061 & 0.027    & 0.976  & 0.997  & 0.999& \checkmark\\
        \hline
        \multicolumn{9}{c}{\textit{Diffusion-Based (without extra training data)}}\\
                \hline
                DDP \cite{ji2023ddpdiffusionmodeldense}    &  ICCV'23   & 0.329 & 0.094 & 0.040 & 0.921  & 0.990  & 0.999 & $\times$\\
        ECoDepth \cite{patni2024ecodepth}   &CVPR'24& \textbf{0.218} & \textit{0.059} & 0.026 & \textbf{0.978}  & 0.997  & 0.999 & $\times$\\
                \hline\rowcolor{Gray}\underline{\textbf{EDADepth(ours)}}     &ICMLA'24 &\textit{0.223} & 0.061 & \textit{0.026} & \textit{0.977}  & \textbf{0.998}  & \textbf{1.000}& $\times$\\
                \hline
    \end{tabular}
    \label{tab:nyuv2}
\end{table*}

\begin{figure*}[!htbp] 
    \centering
    \vspace{-6pt}
\includegraphics[width=\linewidth]{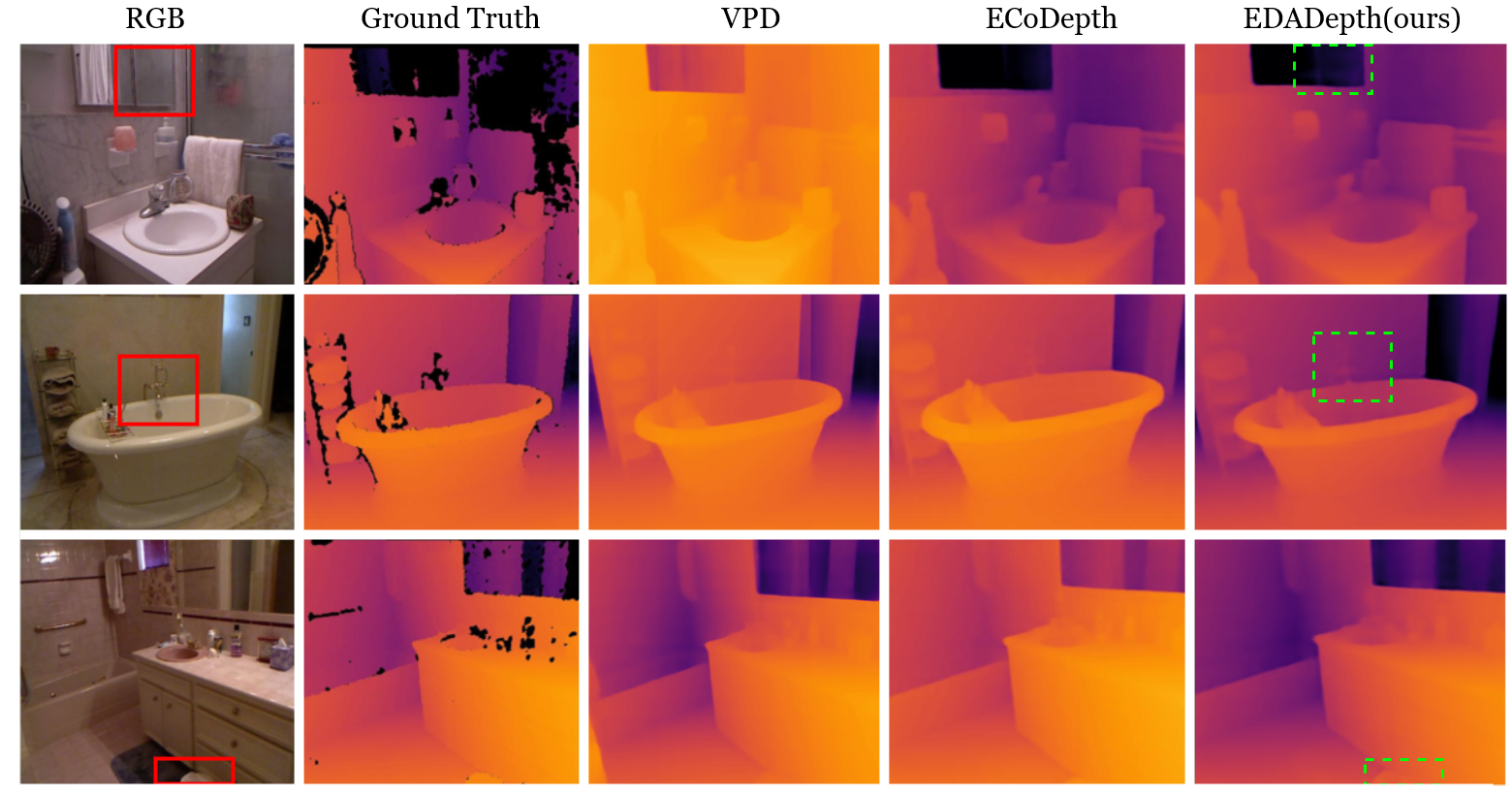} 
    \caption{Comparison of different diffusion-based MDE models with their test samples. The annotated green box denotes the area where the visualization of the output from our model outperforms the visualization of ECoDepth~\cite{patni2024ecodepth}, the current diffusion-based SOTA. Zoom-in for better visibility.} 
    \label{datasets}
\end{figure*}

\begin{figure*}[!htb] 
\centering
\vspace{-6pt}
\includegraphics[width=\linewidth]{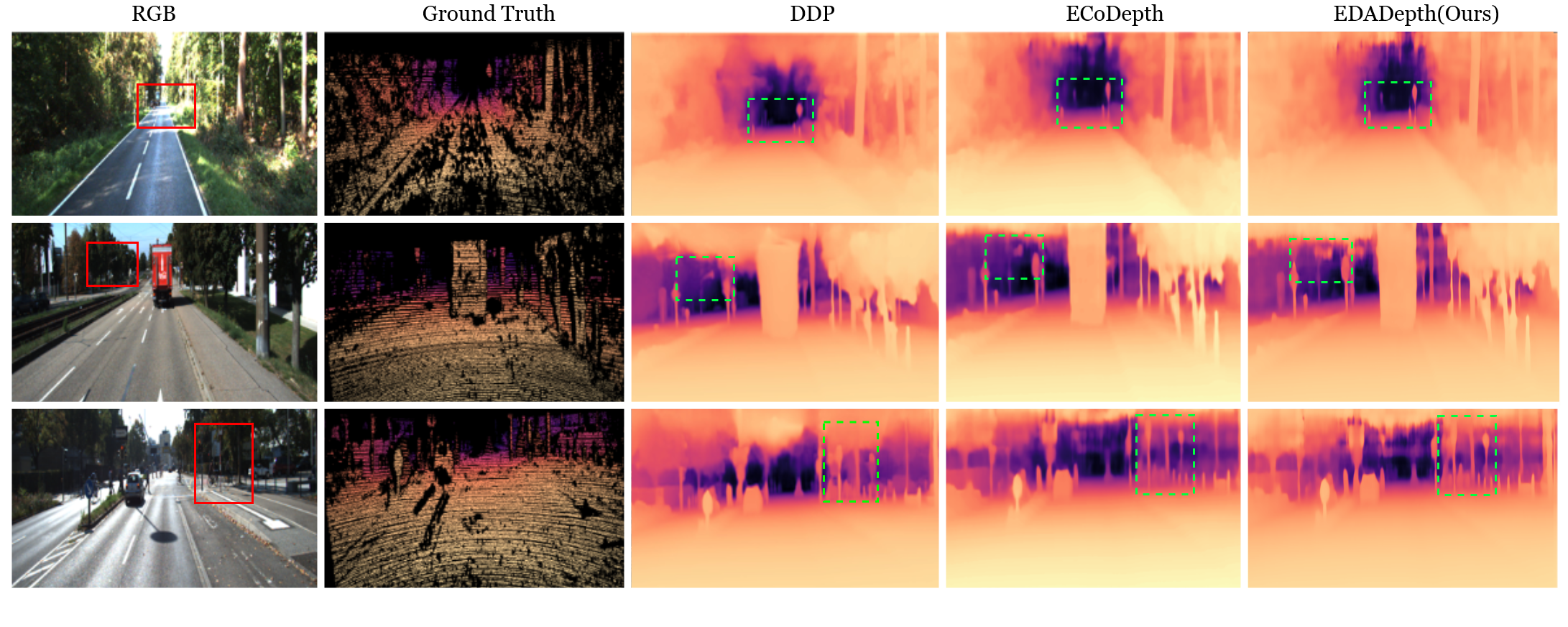} 
\caption{Comparison of various diffusion-based monocular depth estimation (MDE) models on KITTI Eigen-Split \cite{eigen2014depthmappredictionsingle} test samples, all trained without additional data. The annotated green box highlights the area where our model's depth estimation surpasses that of the other models. Zoom-in for better visibility.}
\label{kitticomp}
\end{figure*}

\begin{table*}[htbp]
    \centering
    \caption{Comparision of models on KITTI Eigen-Split \cite{eigen2014depthmappredictionsingle} Dataset. The \textbf{bold metrics} represent SOTA, and the \textit{italic metrics} represent the second-best performance. The row with a light gray fill represents the performance of our model, EDADepth.}
\begin{tabular}{ c|c|c|c|c|c|c|c|c}
        \hline
        \textbf{Method} & 
        \textbf{Venue} & \textbf{RMSE$\downarrow$} & \textbf{REL$\downarrow$} & \textbf{RMSE$_{log}$$\downarrow$} & \textbf{$\delta_{1}$$\uparrow$} & \textbf{$\delta_{2}$$\uparrow$} & \textbf{$\delta_{3}$$\uparrow$} &
        \textbf{extra training data}\\
        \hline
                \multicolumn{9}{c}{\textit{Non-Diffusion-Based}}\\
                \hline
        Eigen et al. \cite{eigen2014depthmappredictionsingle}& NIPS'14        & 6.3041 & 0.203 & 0.282 & 0.702  & 0.898  & 0.967 & $\times$\\
        DORN \cite{fu2018deepordinalregressionnetwork}&  CVPR'18        & 2.727 & 0.072 & 0.120 & 0.932  & 0.984  & 0.994 & $\times$\\
        Yin et al. \cite{yin2020learningrecover3dscene} & CVPR'21 & 3.258 & 0.072 & 0.117 & 0.938  & 0.990  & 0.998 & $\times$\\
        BTS\cite{lee2021bigsmallmultiscalelocal} & Arxiv'19        & 2.756 & 0.059 & 0.096 & 0.885  & 0.978  & 0.994 & $\times$\\
        TransDepth\cite{yang2021transformerbasedattentionnetworkscontinuous} & ICCV'21        & 2.755 & 0.064 & 0.098 & 0.956  & 0.994  & 0.999 & $\times$\\
        AdaBins\cite{Farooq_Bhat_2021}&  CVPR'21        & 2.960 & 0.067 & 0.088 & 0.949  & 0.992  & 0.998 & $\times$\\
        DPT\cite{ranftl2021visiontransformersdenseprediction}& ICCV'21        & 2.573 & 0.060 & 0.092 & 0.959  & 0.995 & 0.996 & \checkmark\\
        P3Depth\cite{patil2022p3depthmonoculardepthestimation}& CVPR'22        & 2.842 & 0.071 & 0.103 & 0.953  & 0.993 & 0.998 & $\times$\\
        NeWCRFs\cite{yuan2022newcrfsneuralwindow} & CVPR'22        & 2.129 & 0.052 & 0.077 & 0.974  & 0.997 & 0.999 & $\times$\\
        DepthFormer\cite{Li_2023} &ArXiv'22         & 2.143 & 0.052 & 0.079 & 0.975 & 0.997 & 0.999 & $\times$\\
        PixelFormer\cite{agarwal2022attentionattentioneverywheremonocular} & WACV'23 & 2.081 & 0.051 & 0.077 & 0.976  & 0.997  & 0.999 & $\times$\\  
        ZoeDepth  \cite{bhat2023zoedepth}       &ArXiv'23& 2.440 & 0.054 & 0.083 & 0.970  & 0.996  & 0.999 & \checkmark\\
        WorDepth \cite{zeng2024wordepthvariationallanguageprior}& CVPR'24& 2.039 & 0.049 & 0.074 & 0.932  & 0.992  & 0.998 & $\times$\\
        MIM \cite{xie2022revealingdarksecretsmasked}&  CVPR'23       & \textbf{1.966} & 0.050 & 0.075 & 0.977  & 0.998  & 1.000 & $\times$\\
                \hline
        \multicolumn{9}{c}{\textit{Diffusion-Based (with extra training data)}}\\
                \hline
               
        Marigold \cite{ke2023repurposing} &CVPR'24 &3.304 &0.099 & 0.138    & 0.916  & 0.987  & 0.996& \checkmark\\
        MetaPrompts \cite{wan2023harnessing} &ArXiv'23 &\textit{1.928} &\textbf{ 0.047} & \textbf{0.071}    & \textbf{0.981}  & \textbf{0.998}  & 1.000& \checkmark\\
        \hline
        \multicolumn{9}{c}{\textit{Diffusion-Based (without extra training data)}}\\
        \hline
         DDP  \cite{ji2023ddpdiffusionmodeldense}  &  ICCV'23   & 2.072 & 0.050 & 0.076 & 0.975  & 0.997  & 0.999 & $\times$\\
        ECoDepth~\cite{patni2024ecodepth}   & CVPR'24 & 2.039 & \textit{0.048} & \textit{0.074} & \textit{0.979}  & 0.998  & 1.000 & $\times$\\

                \hline
        \rowcolor{Gray}\underline{\textbf{EDADepth (ours)}}     & ICMLA'24&2.070 & 0.051 & 0.077 & 0.978  & \textit{0.997}  & \textbf{1.000}& $\times$\\
                \hline
    \end{tabular}
    \label{tab:kitti}
\end{table*}

\section{Experimental Results}
\subsection{Datasets and Evaluation}
The monocular depth estimation (MDE) models are trained and evaluated using the NYU Depth v2 and KITTI data sets. NYU Depth v2 \cite{Silberman:ECCV12} dataset consists of video sequences in indoor scenes with 24,231 2D 640$\times$480 RGB images and the corresponding infrared-based depth maps have a depth range of 0.1 to 10 meters. KITTI Eigen-Split \cite{eigen2014depthmappredictionsingle} dataset consists of video sequences in outdoor scenes containing 23,158 2D 1240$\times$375 RGB images and its corresponding infrared-based depth maps having a depth range of 0.1 to 80 meters. The results are measured using RMSE (root mean squared error) and REL (absolute relative error) as the primary evaluation metrics. The average error $log_{10}$ between the predicted depth $a$ and the actual depth $d$, and the threshold accuracy $\delta_{n}$ which measures the $\%$ of pixels that satisfy the max$(a_{i}/d_{i},d_{i}/a_{i}) < 1.25^{n}$, where $ n=1,2,3$, are other common metrics as in Tables \ref{tab:nyuv2}, \ref{tab:kitti}. We also provide qualitative results in figures \ref{datasets},\ref{kitticomp}. 

\subsection{Implementaton Details}
Our proposed EDADepth model uses the HuggingFace Stable-Diffusion-v1-5 checkpoint as the U-Net diffusion backbone. For super-resolution, we utilized a Swin2SR model from HuggingFace to upscale the resolution by 2x. Additionally, we incorporated the BEiT-based model for semantic segmentation. We trained using 8 NVIDIA H100 GPUs~\cite{10070122} for 25 epochs, with a total batch size of 32. Our model did not use additional training data, but relied solely on the original NYUv2Depth \cite{Silberman:ECCV12} and KITTI Eigen-Split \cite{eigen2014depthmappredictionsingle} datasets.

\subsection{Quantitative Results}
As shown in Table \ref{tab:nyuv2} for the NYUv2 test set, among the stable diffusion-based models (both with and without additional training data), our model achieved the second-best results in metrics such as RMSE, log10 and $\delta_1$. Our model achieved SOTA results for $\delta_2$ and $\delta_3$ among diffusion-based models. For the KITTI Eigen-Split dataset, Table \ref{tab:kitti} shows that our model achieved SOTA for $\delta_3$, indicating better visualizations through precise depth estimation than other diffusion-based models. These results support our model's ability to generate precise depth maps and improve visualization in outdoor datasets.

\subsection{Qualitative Results}
Figure \ref{datasets} compares our model with recent SOTA models trained and evaluated on the NYUv2 Test Dataset. The three rows showcase our model's superior depth estimation. The red boxes in the first column highlight regions of interest in RGB images, while the green boxes in the last column show where our model outperforms existing diffusion-based methods. In the top row, a green dotted box marks a mirror accurately captured by our model. In the Our model correctly identifies a "Faucet" in the second row depth map. Similarly, in the bottom row, the green dotted box highlights a "Towel" accurately included by our model. 

Figure \ref{kitticomp} illustrates our model performance in the KITTI eigen-split data set. We achieved results comparable to SOTA diffusion-based models. As demonstrated with the NYUv2 dataset, our model excels in depth estimation. The red boxes in the first column highlight objects where our model captures intricate details with greater precision. In the upper row, a red box indicates a street sign that is not visible in the original image because of its resolution, but our model accurately identifies it (last column). In the second row, a red box highlights a traffic light pole that our model captures with greater precision. Similarly, in the bottom row, our model more accurately represents street signs than other models.

\section{Ablation Study}
\textbf{Extraction of text embeddings:} We experimented using SOTA backbones to extract semantic context and generate text embedding vectors on the NYUv2 dataset. As shown in Table \ref{tab:ablation}, the BEiT-Base model backbone outperforms other models by providing better results on the given metrics.   

\begin{table}[htbp]
\centering
\vspace{-5pt}
\caption{Comparison of various models performing different tasks for obtaining semantic context to generate text embeddings. BEiT-Base performs better for the provided metrics.
\textbf{ImgC: Image Classification}, \textbf{SSeg: Semantic Segmentation}}
\begin{tabular}{lcccc}
\toprule
\textbf{Model} & \textbf{Task} & \textbf{RMSE$\downarrow$} & \textbf{REL$\downarrow$}& \textbf{$log_{10}$$\downarrow$}\\
\midrule
SwinV2-Base \cite{liu2021swinv2}& ImgC & $0.227$ & $0.062$ & $0.027$\\
SegFormer-Base \cite{xie2021segformersimpleefficientdesign} & SSeg & $0.225$ & $0.062$ & $0.027$\\
BEiT-Base \cite{bao2022beitbertpretrainingimage}& SSeg & \textbf{0.223} & \textbf{0.061} & \textbf{0.026}\\
\bottomrule
\vspace{-15pt}
\end{tabular}
\label{tab:ablation}
\end{table}

\section{Conclusion}
In this paper, we proposed \textbf{EDADepth}, a novel method for monocular depth estimation using the enhanced super-resolution data enhancement technique. Firstly, we focused on superresolving the input data using a pre-trained Swin2SR model to improve the extraction of textual embeddings and the denoising process in the U-Net framework. Secondly, we employed a pre-trained BEiT Semantic Segmentation model to generate text embeddings to capture the semantic context from the input images. Third, we introduce the use of BLIP-2 as a tokenizer. Finally, we conducted extensive experiments on the NYUDepthv2 and KITTI Eigen-split datasets, demonstrating the effectiveness of our method. Our quantitative results show that our model achieves RMSE and REL values comparable to the current SOTA models while achieving SOTA on $\delta_{3}$ values. From our qualitative results, it is evident that EDADepth competes closely with diffusion-based (both with and without using extra training data) SOTA models, particularly in enhancing the generation of visually detailed depth estimation.

{\small
\bibliography{egbib}
}

\end{document}